\documentclass[10pt,twocolumn,letterpaper]{article}

\usepackage{cvpr}              

\usepackage{graphicx}
\usepackage{amsmath}
\usepackage{amssymb}
\usepackage{booktabs}
\usepackage{makecell}
\usepackage{enumitem}

\usepackage[accsupp]{axessibility}

\usepackage[pagebackref,breaklinks,colorlinks]{hyperref}

\usepackage{tabularx}
\usepackage{multirow}
\usepackage[table,xcdraw]{xcolor}
\usepackage[capitalize]{cleveref}

\definecolor{myred}{RGB}{255, 80, 80} 
\hypersetup{
	colorlinks=true,    
	urlcolor=myred,
}

\graphicspath{{./images/}}

\crefname{section}{Sec.}{Secs.}
\Crefname{section}{Section}{Sections}
\Crefname{table}{Table}{Tables}
\crefname{table}{Tab.}{Tabs.}

\def\confName{CVPR}
\def\confYear{2023}

\begin{document}

\title{MIANet: Aggregating Unbiased Instance and General Information for Few-Shot Semantic Segmentation}


\author{Yong Yang$^{1}$ \quad Qiong Chen$^{1}$\thanks{Corresponding author (csqchen@scut.edu.cn).} \quad  Yuan Feng$^{1, 2}$ \quad Tianlin Huang$^{1}$ \\
	 $^1$School of Computer Science and Engineering, South China University of Technology\\
	 $^2$Guangdong Provincial Key Laboratory of Artificial Intelligence in Medical Image Analysis and Application
}

\maketitle

\begin{abstract}
 Existing few-shot segmentation methods are based on the meta-learning strategy and extract instance knowledge from a support set and then apply the knowledge to segment target objects in a query set. However, the extracted knowledge is insufficient to cope with the variable intra-class differences since the knowledge is obtained from a few samples in the support set. To address the problem, we propose a multi-information aggregation network (MIANet) that effectively leverages the general knowledge, i.e., semantic word embeddings, and instance information for accurate segmentation. Specifically, in MIANet, a general information module (GIM) is proposed to extract a general class prototype from word embeddings as a supplement to instance information. To this end, we design a triplet loss that treats the general class prototype as an anchor and samples positive-negative pairs from local features in the support set. The calculated triplet loss can transfer semantic similarities among language identities from a word embedding space to a visual representation space. To alleviate the model biasing towards the seen training classes and to obtain multi-scale information, we then introduce a non-parametric hierarchical prior module (HPM) to generate unbiased instance-level information via calculating the pixel-level similarity between the support and query image features. Finally, an information fusion module (IFM) combines the general and instance information to make predictions for the query image. Extensive experiments on PASCAL-5$^i$ and COCO-20$^i$ show that MIANet yields superior performance and set a new state-of-the-art. Code is available at \href{https://github.com/Aldrich2y/MIANet}{github.com/Aldrich2y/MIANet}.
\end{abstract}

\section{Introduction}
\begin{figure}[htbp]
	\centering
	\includegraphics[width=1\columnwidth,height=0.8\linewidth]{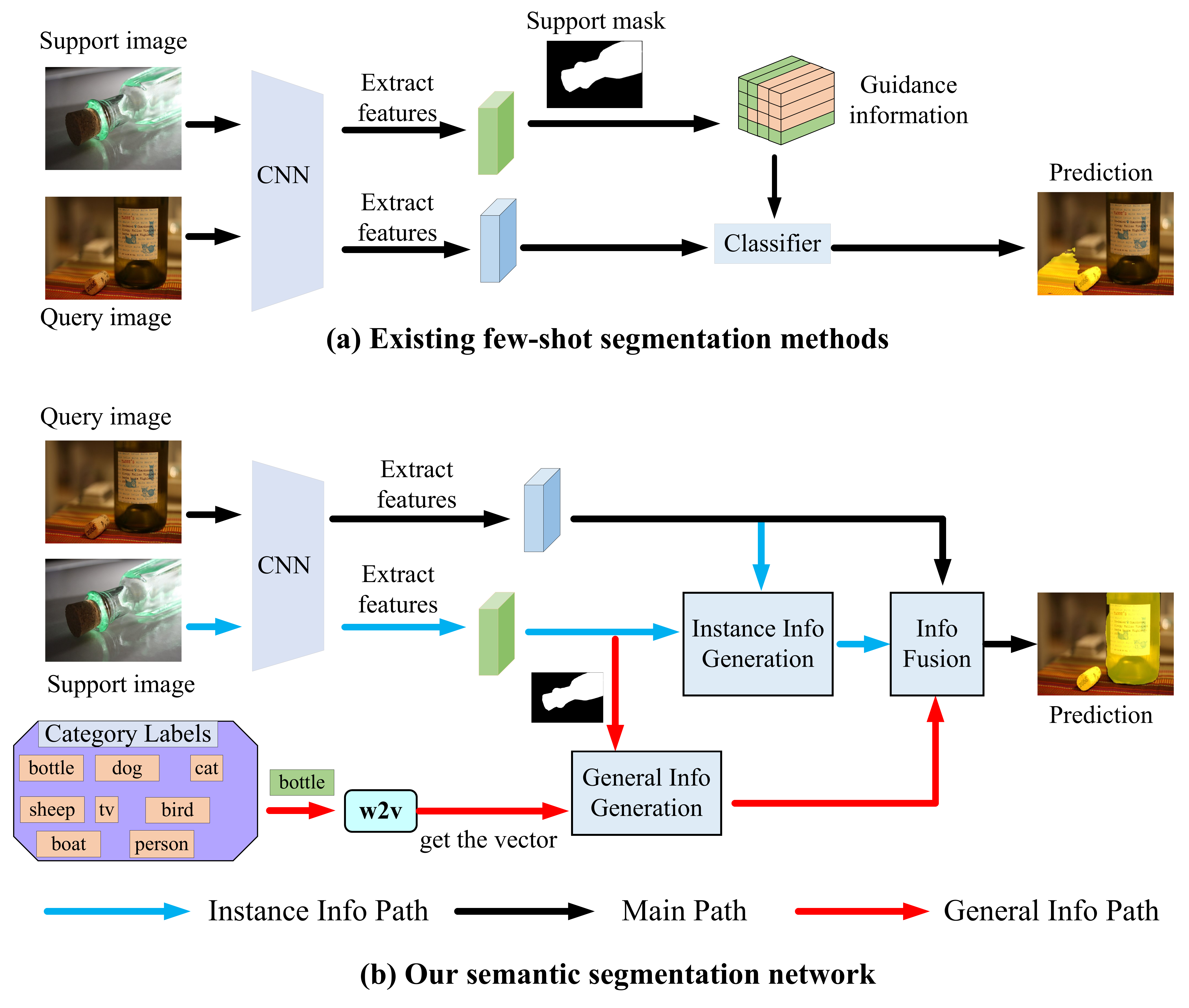}
	\caption{Comparison between (a) existing FSS methods and (b) proposed MIANet. (a) Existing methods extract instance-level knowledge from the support images, which is not able to cope with large intra-class variation. (b) our MIANet extracts instance-level knowledge from the support images and obtains general class information from word embeddings. These two types of information benefit the final segmentation.}
	\label{figure1}
\end{figure}
The challenge of few-shot semantic segmentation (FSS) is how to effectively use one or five labeled samples to segment a novel class. Existing few-shot segmentation methods \cite{pfenet, panet, canet,cmn} adopt the metric-based meta-learning strategy \cite{meta-learning, oslsm}. The strategy is typically composed of two stages: meta-training and meta-testing. In the meta-training stage, models are trained by plenty of independent few-shot segmentation tasks. In meta-testing, models can thus quickly adapt and extrapolate to new few-shot tasks of unseen classes and segment the novel categories since each training task involves a different seen class.\par
\begin{figure}[htbp]
	\centering
	\includegraphics[width=1\columnwidth,height=0.6\linewidth]{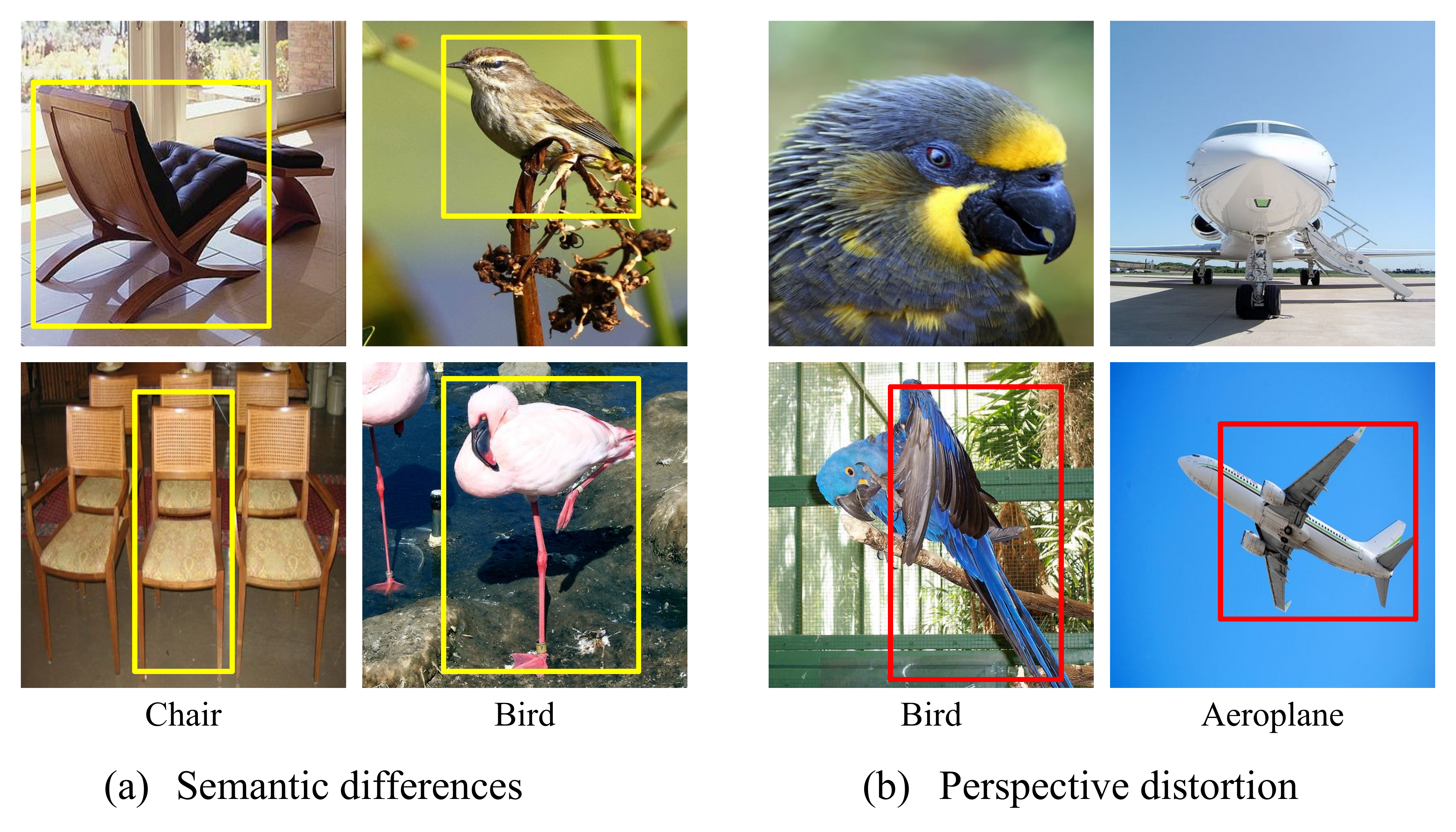}
	\caption{We define two types of intra-class variation. (a) The object in each column has the same semantic label but belongs to different fine-grained categories. (b) The object belonging to the same category differs greatly in appearance due to the existence of perspective distortion.}
	\label{figure2}
\end{figure}

As shown in Figure \ref{figure2}, natural images of same categories have semantic differences and perspective distortion, which leads to intra-class differences. Current FSS approaches segment a query image by matching the guidance information from the support set with the query features (Figure \ref{figure1} (a)). Unfortunately, the correlation between the support image and the query image is not enough to support the matching strategy in some support-query pairs due to the diversity of intra-class differences, which affects the generalization performance of the models. On the other hand, modules with numerous learnable parameters are devised by FSS methods to better use the limited instance information. And lots of few-shot segmentation tasks of seen classes are used to train the models in the meta-training stage. Although current methods freeze the backbone, the rest parameters will inevitably fit the feature distribution of the training data and make the trained models misclassify the seen training class to the unseen testing class.

To address the above issues, a multi-information aggregation network is proposed for accurate segmentation. Specifically, we first design a general information module (GIM) to produce a general class prototype by leveraging class-based word embeddings. This prototype represents general information for the class, which is beyond the support information and can supplement some missing class information due to intra-class differences. As shown in Figure \ref{figure1} (b), the semantic word vectors for each class can be obtained by a pre-trained language model, i.e., \textit{word2vec}. Then, GIM takes the word vector and a support prototype as input to get the general prototype. Next, a well-designed triplet loss \cite{facenet} is applied to achieve the alignment between the semantic prototype and the visual features. The triplet loss extracts positive-negative pairs from local features which distinguishes our method from other improved triplets \cite{improved,quadruplet,indefense}. The semantic similarity between the word embeddings in a word embedding space can therefore be transferred to a visual embedding space. Finally, the projected prototype is supplemented into the main branch as the general information of the category for information fusion to alleviate the intra-class variance problem.\par
Moreover, to capture the instance-level details and alleviate the model biasing towards the seen classes, we propose a non-parametric hierarchical prior module (HPM). HPM works in two aspects. (1) HPM is class-agnostic since it does not require training. (2) HPM can generate hierarchical activation maps for the query image by digging out the relationship between high-level features for accurate segmentation of unseen classes. In addition, we build information channels between different scales to preserve discriminative information in query features. Finally, the unbiased instance-level information and the general information are aggregated by an information fusion module (IFM) to segment the query image. Our main contributions are summarized as follows: 
\begin{itemize}[itemsep=1pt,topsep=0pt,parsep=0pt]
	\item [(1)] We propose a multi-information aggregation network (MIANet) to aggregate general information and unbiased instance-level information for accurate segmentation. 
	\item [(2)] To the best of our knowledge, this is the first time to use word embeddings in FSS, and we design a general information module (GIM) to obtain the general class information from word embeddings for each class. The module is optimized through a well-designed triplet loss and can provide general class information to alleviate intra-class differences.
	\item [(3)] A non-parametric hierarchical prior module (HPM) is proposed to supply MIANet with unbiased instance-level segmentation knowledge, which provides the prior information of the query image on multi-scales and alleviates the bias problem in testing.
	\item [(4)] Our MIANet achieves state-of-the-art results on two few-shot segmentation benchmarks, i.e., PASCAL-5$^i$ and COCO-20$^i$. Extensive experiments validate the effectiveness of each component in our MIANet.
\end{itemize}

\section{Related work}
\noindent\textbf{Few-Shot Semantic Segmentation.} Few-shot semantic segmentation (FSS) is proposed to address the dependence of semantic segmentation models on a large amount of annotated data. Current FSS methods are based on metric-based meta-learning and can be largely grouped into two types: prototype-based methods \cite{panet,mlc,ppnet,sgone, pl, apla} and parameter-based methods \cite{sagnn, cytrans, bam, crnet,self,metaclass}. The prototype-based methods use a non-parametric metric tool, e.g., cosine similarity or euclidean distance, to calculate segmentation guidance. And non-parametric metric tools alleviate overfitting. The parameter-based FSS methods employ learnable metric tools to explore the relationship between the support and query features. For instance, BAM \cite{bam} proposes a base learner to avoid the interference of base classes in testing and achieve the state-of-the-art performance. Current methods can effectively segment the target area of novel classes when samples of the classes are limited. However, these methods only extract instance knowledge from the limited support set, and cannot segment some support-query pairs with large intra-class differences as detailed in Figure \ref{figure2}. For this problem, we propose a multi-information aggregation network, which extracts instance information and learns general class prototypes from word embeddings to alleviate the intra-class differences.\par

\vspace{1mm}

\noindent\textbf{Intra-Class Differences.} The intra-class differences problem is a key factor affecting the performance of the few-shot segmentation. Previous methods try to mine more support information to alleviate this issue. \cite{cwt} dynamically transforms a classifier trained on the support set to each query image. \cite{ipmt, ssp} produce a pseudo query mask based on the support information to capture more self-attention information of the query image. But the performance gain is restricted since the support set is limited. In zero-shot learning (ZSL), semantic information is used to generate visual features for unseen classes \cite{zs3net,cagnet,zsl1,zsl2, zsl3}, so that the models recognize the unseen classes. The achievement in ZSL demonstrates that word embeddings contain the general semantic information of categories, which inspires us to integrate class-based semantic information \cite{w2v, ft} to supplement the missing information when the features in the support set and in the query set don't match.

\section{Methodology}
\subsection{Problem Definition}
We define two datasets, $D_{train}$  and $D_{test}$, with the category set $C_{train}$ and $C_{test}$ respectively, where $C_{train} \cap C_{test}= \emptyset $. The model trained on $D_{train}$ is directly transferred to evaluate on $D_{test}$ for testing. Besides, each category $c \in {C_{train} \cup C_{test}  }$ is mapped through the word embedding to a vector representation $W[c] \in R^d$, where d is the dimension of $W[c]$. In line with previous works \cite{pfenet}, we train the model in an episode manner. Each episode contains a support set $S$, a query set $Q$ and a word embedding map $W$. Under the K-shot setting, each support set $S=\left\{X_s^i, M_s^i\right\}_{i=1}^K$, includes K support images $X_s$ and corresponding masks $M_s$, and each query set $Q=\left\{X_q, M_q\right\}$, includes a query image $X_q$ and a corresponding mask $M_q$. The training set $D_{train}$ and test set $D_{test}$ are represented by $D_{train}=\left\{(S_i,Q_i,W )\right\}_{i=1}^{N_{train}}$ and $ D_{test}=\left\{(S_i,Q_i,W )\right\}_{i=1}^{N_{test}} $, where $N_{train}$ and $N_{test}$ is the number of episodes for training and test set. During training, the support masks $M_s$ and query masks $M_q$ are available, and the $M_q$ is not accessible during testing.

\subsection{Method Overview}
\begin{figure*}[htbp]
	\centering
	\includegraphics[width=2\columnwidth,height=0.4\linewidth]{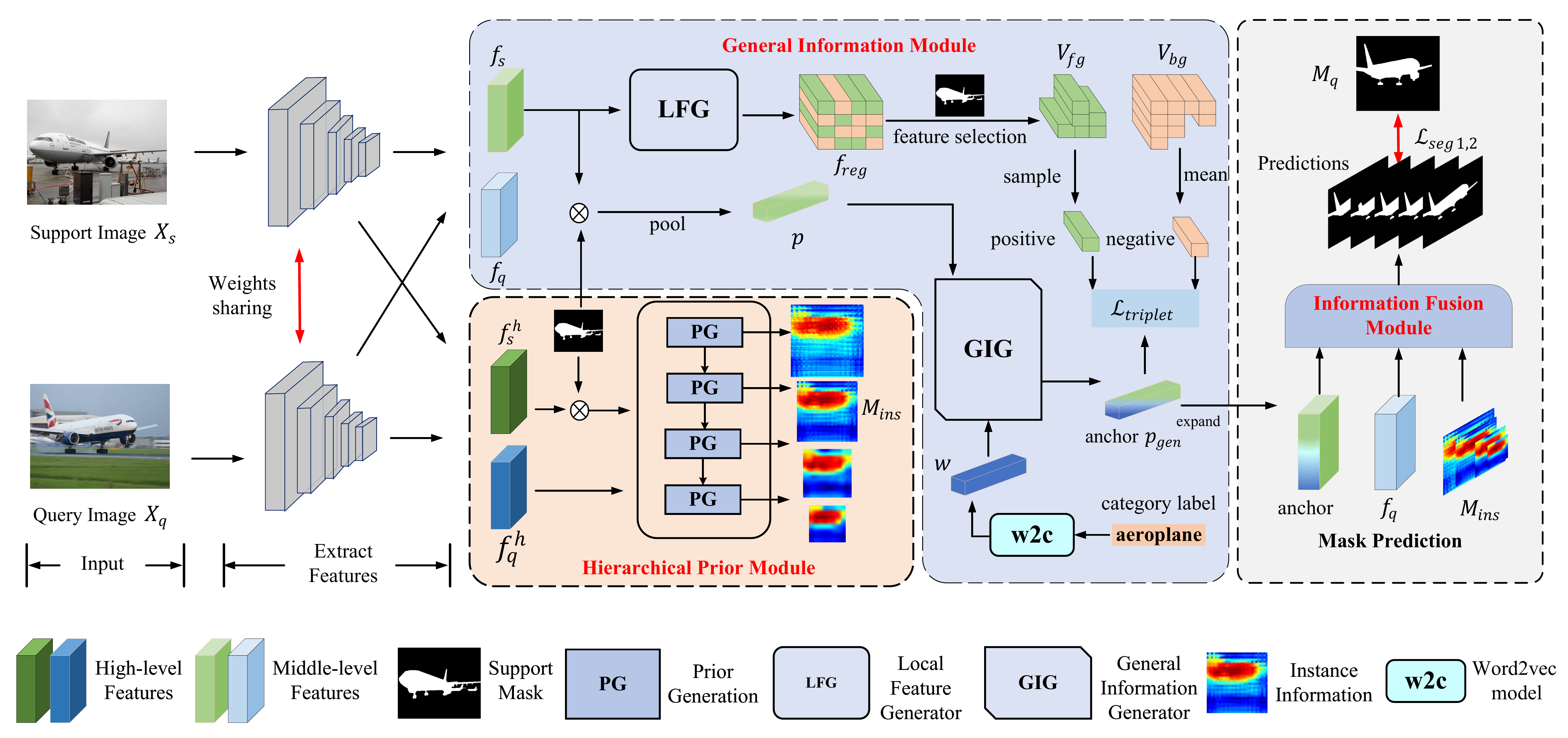}
	\caption{The overall architecture of our proposed multi-information aggregation network.}
	\label{figure3}
\end{figure*}

As shown in Figure \ref{figure3}, our multi-information aggregation network includes three modules, i.e., hierarchical prior module (HPM), general information module (GIM), and information fusion module (IFM). Specifically, given the support and query images $X_s$ and $X_q$, a common backbone with shared weights is used to extract both middle-level \cite{canet} and high-level features \cite{pfenet}. We then employ HPM whose task is to produce unbiased instance-level information $M_{ins}$ of the query image by using labeled support instances. Meanwhile, GIM is introduced to generate general class information which aims to make up for the insufficiency of instance information. At last, we pass the instance information and general information to an information fusion module to aggregate into the final guidance information and then make predictions for the query image.

\subsection{Hierarchical Prior Module}
Few-shot semantic segmentation models are trained on labeled data of seen classes, which makes it inclined for trained models to misjudge seen training categories as unseen target categories. Moreover, current approaches usually resort to well-designed modules with numerous learnable parameters in order to maximize the use of limited support information. Inspired by \cite{pfenet}, we propose a non-parametric hierarchical prior module (HPM) to capture the unbiased instance information from a few labeled samples in an efficient way. HPM leverages the high-level features (e.g., layer 4 of ResNet50) from the support set and query set to generate prior information, which is a rough localization map of the target object in the query image. Moreover, we compute prior information at multiple different scales that provide rich guidance for objects of varying sizes and shapes. In order to avoid the loss of discriminative information when the query features are extended to different scales, we establish information channels between different scales.\par

Specifically, HPM takes as input the high-level support features $f_s^h\in R^{c\times h \times w}$, the corresponding binary mask $M_s \in R^{H \times W}$, and the high-level query features $f_q^h \in R^{c\times h \times w}$, where c is the channel dimension, h (H), w (W) are the height and width of the features and the mask. Empirically \cite{pfenet}, we define the instance-level information as $M_{ins}=\left\{m_{ins}^i \right\}_{i=1}^4$, $m_{ins}^i \in R^{c \times h_i \times w_i}$, and $ h_i>h_j,w_i>w_j$, when $i <j$, $h_1=h, w_1=w$. \par
To obtain the $m_{ins}^1$, we first filter out the background elements in the support features via
\begin{equation}
f_s^h = f_s^h \otimes \mathcal{I}(M_s, f_s^h)
\label{eqn1}
\end{equation}
where $ \mathcal{I}(M_s,f_s^h)$ down- or up-samples the $M_s$ to a spatial size as the $f_s^h$ by interpolation, $\otimes$ means the Hadamard product. Next, we reshape the $f_s^h$ and $f_q^h$ to a size of ($c\times hw$). The pixel-wise cosine similarity $A_q$ between $f_s^h$ and $f_q^h$ is calculated as 
\begin{equation}
A_q = \frac{(f_q^h)^Tf_s^h}{\vert \vert f_q^h \vert \vert \  \vert \vert f_s^h \vert\vert} \in R^{h_1w_1\times h_1w_1}
\label{eqn2}
\end{equation}
We then take the mean similarity in the support (second) dimension as the activation value and pass the $A_q$ into a min-max normalization ($\mathcal{F}_{norm}$) to get the $m_{ins}^1$.
\begin{equation}
m_{ins}^1 = \mathcal{F}_{norm}(mean(A_q)) \in R^{h_1\times w_1}
\label{eqn3}
\end{equation}

In order to extend to the next scale, i.e., $ (h_2,w_2) $, the pooling operation is needed to down-sample the $f_q^h$. We use the weighted average pooling to add information channels between different scales since discriminative details are prone to be ignored by the average pooling
\begin{equation}
f_q^h = \mathcal{F}_{pool}(f_q^h \otimes m_{ins}^1) \in R^{c\times h_2 \times w_2}
\label{eqn6}
\end{equation}
where $\mathcal{F}_{pool}$ is the average pooling. Then the high-level support features in the next stage can be computed by
\begin{equation}
f_s^h = \mathcal{I}(f_s^h, f_q^h) \in R^{c \times h_2 \times w_2}
\label{eqn7}
\end{equation}
\par
Finally, prior information $m_{ins}^2$ can be obtained by using equation \ref{eqn1} - \ref{eqn3}, and $\left\{m_{ins}^i\right\}_{i=1}^4$ can be calculated after four stages.

\subsection{General Information Module}
One of the main challenges of few-shot semantic segmentation is the intra-class differences as shown in Figure \ref{figure2}. Current methods aim to address this problem by thoroughly excavating the relationship between instance samples and the query image, i.e., digging out the instance-level information. But this can only solve some highly correlated support-query pairs. For instance, in the case of Figure \ref{figure2} (1st and 2nd columns), objects in the support image and the query image have similar local features despite belonging to different fine-grained categories, such as the legs of the chair, the feathers, and the body of the bird. But in Figure \ref{figure2} (b), due to the existence of perspective distortion, some local features (the part in the red box) are lost, and it is difficult for the model to segment the query image according to the incomplete support sample.\par
To counter this, a general information module (GIM) is used to extract language information from word embeddings to generate a general class prototype, and a triplet loss is designed to optimize this module. GIM contains two components: general information generator (GIG) and local feature generator (LFG). GIG takes the foreground prototype obtained from the support set and the category semantic vector obtained from the semantic label as input, and generates a general class prototype. LFG takes the mid-level support features as input and generates region-related local features to collect positive-negative pairs to form triplets.\par
Specifically, we input the category word (e.g., \textit{aeroplane}) to the pre-trained \textit{word2vec} to obtain a vector representation $w\in R^{1\times d}$. 
\begin{equation}
w = \mathcal{F}_{word2vec}(word)
\label{eqn9}
\end{equation}
where $\mathcal{F}_{word2vec}(.)$ represents generating vector representation from the word embeddings according to $word$.
\par 
Next, masked average pooling is applied on the support features $f_s \in R^{c\times h\times w}$ to get a foreground class prototype $p \in R^{1\times c}$ as 
\begin{equation}
p = \mathcal{F}_{pool}(f_s \otimes \mathcal{I}(M_s, f_s))
\label{eqn8}
\end{equation}

Then, we input the foreground class prototype $p$ and the word vector $w$ into GIG to produce a general class prototype $p_{gen}\in R^{1\times c}$
\begin{equation}
p_{gen} = \mathcal{F}_{GIG}(w \oplus p)
\label{eqn}
\end{equation}
where $\oplus$ is the concatenation operation in channel dimension, $\mathcal{F}_{GIG}(.)$ means producing the general information, GIG consists of two fully connected layers.\par

The obtained prototype $p_{gen}$ represents the general and complete information for a specific category, which is expected to distinguish whether a local feature belongs to the category. To achieve this, we set $p_{gen}$ as the \textit{anchor}, and then sample pairs of \textit{positive} and \textit{negative} from local features to calculate the triplet loss. Different from pixel-level features, local features are region-related and represent part of the semantic information of categories, such as the tail, head, torso, and other features. We design a local feature generator (LFG) which consists of three convolutional blocks and reduces the size of the support features by a factor of 4 to obtain regional features. A regional vector $v\in R^{1\times c}$ in the regional features $f_{reg}$ can represent an area in the original image, i.e., a local feature representation.
\begin{equation}
f_{reg} = \mathcal{F}_{reshape}^{hw\times c}(\mathcal{F}_{LFG}(f_s)) \in R^{hw\times c}
\label{eqn11}
\end{equation}
where $\mathcal{F}_{LFG}(.)$ indicates generating the local information, and $\mathcal{F}_{reshape}^{hw\times c}(.)$ means reshaping the input to a spatial size of $(hw\times c)$. We then use support mask $M_s\in R^{H\times W}$ for feature selection, which separates the foreground and background regional vectors into two different sets, i.e., $V_{fg}=\left\{v_{fg}^i\right\}_{i=1}^{n_1}, V_{bg}=\left\{v_{bg}^i\right\}_{i=1}^{n_2}, v_{bg},v_{fg}\in R^{1\times c}, n1+n2=hw$.

\begin{equation}
\hat{M_s} = \mathcal{F}_{reshape}^{hw\times 1}(\mathcal{I}(M_s, f_{reg})) \in R^{hw\times 1}  
\end{equation}

\vspace{-5.5mm}
\begin{equation}
V_{fg} = \mathcal{F}_{index}(\hat{M}_s^k == 1, f_{reg}^k) \ \ k \in \left\{1,2,...,hw\right\}
\end{equation}

\vspace{-5.5mm}
\begin{equation}
V_{bg} = \mathcal{F}_{index}(\hat{M}_s^k == 0, f_{reg}^k) \ \ k \in \left\{1,2,...,hw\right\}
\end{equation}
where $\mathcal{F}_{index}(\hat{M}_s^k, f_{reg}^k)$ indicates that when $\hat{M}_s^k$ is 1, add the corresponding vector $f_{reg}^k$ to $V_{fg}$, otherwise, add it to $V_{bg}$. Next, we average the $V_{bg}$ to get negative sample since the elements in the background of the support images are very complex and are hard to use \cite{panet}.
\begin{equation}
negative = \frac{\sum_{i}^{n_2} (v_{bg}^i)}{n_2}, \ \  v_{bg}^i \in V_{bg}
\end{equation}
\par
The positive samples are the foreground regional vectors in $V_{fg}$. Similar to \cite{indefense}, we calculate the hardest sample, which has the farthest distance from the \textit{anchor}, to obtain the positive vector for better optimization.
\begin{equation}
positive = \mathop{\arg\max}\limits_{v_{fg}^i}(\mathcal{F}_d(p_{gen}, v_{fg}^i)), \ \  v_{fg}^i \in V_{fg}
\end{equation}
where $\mathcal{F}_d$ is the $l_2$ distance function. The triplet loss $\mathcal{L}_{triplet}$ is
\begin{equation}
\label{tripletloss}
\begin{aligned}
\mathcal{L}_{triplet}=\max(\mathcal{F}_d(p_{gen}, positive) + margin \\ - \mathcal{F}_d(p_{gen}, negative), 0)
\end{aligned}
\end{equation}
where margin is a fixed value (0.5) to keep negative samples far apart. \par
By calculating the distance among triplets (anchor, foreground local features, background local features), the semantic information of the anchor and the visual information of local features are aligned, and the relationship among different word vectors can also be converted to visual embedding space to provide additional general information to alleviate the intra-class differences even some features are lost due to perspective distortion in Figure \ref{figure2} (b). In addition, the triplet loss encourages the GIG to learn better general prototypes (\textit{anchor}) to distinguish fine-grained local features (\textit{positive}) of the same category from background features (\textit{negative}).

\subsection{Prediction and Training Loss}
The instance-level information $M_{ins}$ and general information $p_{gen}$ are aggregated as guidance information through the information fusion module (IFM) to supervise the segmentation of query images. In order to seek more contextual cues, we utilize the FEM \cite{pfenet} structure as our information fusion module. As shown in Figure \ref{figure3}, the mid-level query feature $f_q$, instance information $M_{ins}$ and general class information $p_{gen}$ are input to IFM. The $f_q$ and $p_{gen}$ are first expanded to four scales$\left\{p_{gen}^i\right\}_{i=1}^4$,$\left\{f_q^i \right\}_{i=1}^4$, according to the size of $M_{ins}$.
\begin{equation}
f_q^i = \mathcal{I}(f_q, m_{ins}^i) \in R^{c\times h_i\times w_i}, i=\left\{1,2,3,4\right\}
\end{equation}
\vspace{-5mm}
\begin{equation}
p_{gen}^i=\mathcal{F}_{expand}(\mathcal{I}(p_{gen}, m_{ins}^i)) \in R^{c\times h_i\times w_i}
\end{equation}
where $\mathcal{F}_{expand}(.)$ means expanding the input in channel dimension. We then input the $\left\{m_{ins}^i\right\}_{i=1}^4$,$\left\{p_{gen}^i\right\}_{i=1}^4$,$\left\{f_q^i \right\}_{i=1}^4$ to FEM to compute the binary intermediate predictions $Y_{inter} = \left\{y^i\right\}_{i=1}^4$ and final prediction $Y$, where $Y, y^i\in R^{H \times W}$.\par
The training loss has two parts, namely the segmentation loss and the triplet loss. The segmentation loss is calculated using multiple cross-entropy functions, with $L_{seg1}$ on the intermediate predictions $Y_{inter}$ and $L_{seg2}$ on the final prediction $Y$. The triplet loss is computed from the hardest triplet, as shown in equation \ref{tripletloss}. The final loss is
\begin{equation}
\mathcal{L} = \mathcal{L}_{seg1} + \mathcal{L}_{seg2} + \mathcal{L}_{triplet}
\end{equation}

\subsection{Extending to K-Shot Setting}
The above discussions focus on the 1-shot setting. For the K-shot setting, K support samples $\left\{X_s^i, M_s^i\right\}_{i=1}^K$ are available. Our method can be easily extended to the K-shot setting. First, K sets of instance information $\left\{M_{ins}^i\right\}_{i=1}^K$ are computed respectively using the K samples. We then average the instance information separately at different scales to get $\hat{M}_{ins}=\left\{\hat{m}_{ins}^j\right\}_{j=1}^4$ for the subsequent process. 

\begin{equation}
\hat{m}_{ins}^j = \frac{1}{K} \sum_{i=1}^{K}m_{ins}^{j;i}
\end{equation}

In addition, the K prototypes obtained by Equation \ref{eqn8} are also averaged. Finally, the local feature $ f_{reg}$ will be obtained from the union of K support features through equation \ref{eqn11}. 

\section{Experiments}

\subsection{ Experimental Settings}
\noindent\textbf{Datasets.} Experiments are conducted on two commonly used few-shot segmentation datasets, PASCAL-5$^i$ and COCO-20$^i$, to evaluate our method. PASCAL-5$^i$ is created from PASCAL VOC 2012 \cite{PASCAL} with additional annotations from SBD \cite{sbd}. The total 20 classes in the dataset are evenly divided into 4 folds $i \in \left\{0,1,2,3\right\}$ and each fold contains 5 classes. The COCO-20$^i$ is proposed by \cite{fwb}, which is conducted from MSCOCO \cite{COCO}. Similar to PASCAL-5$^i$, 80 classes in COCO-20$^i$ are partitioned into 4 folds and each fold contains 20 classes.\par

\noindent\textbf{Metric and Evaluation.} We follow the previous methods and adopt the mean intersection-over-union (mIoU) and foreground-background IoU (FB-IoU) as the evaluation metrics. The FB-IoU results are listed in the supplementary material. During testing, we follow the settings of PFENet to make the experimental results more accurate. Specifically, five different random seeds are set for five tests in each experiment. In each test, 1000 and 5000 support-query pairs are sampled for PASCAL-5$^i$ and COCO-20$^i$ respectively. We then average the results of five tests for each experiment. \par

\noindent\textbf{Implementation Details.} Following \cite{bam,cwt}, we first train the PSPNet \cite{ppnet} to obtain a feature extractor (backbone) based on the seen training classes for each fold, i.e., 16/61 training classes (including background) for PASCAL-5$^i$/COCO-20$^i$. Next, we fix the parameters of the trained feature extractor and use a meta-learning strategy to train the remaining structures. These structures are optimized using the SGD optimizer, trained for 200 epochs on PASCAL-5$^i$ and 50 on COCO-20$^i$. The learning rate and batch size are 5e-3 and 4, respectively. And we use the \textit{word2vec} model learned on google news to obtain d (300) dimensional word vector representations. The word embeddings of categories that contain multiple words are obtained by averaging the embeddings of each individual word. \par

\noindent\textbf{Baseline.} As shown in Figure \ref{figure3}, we first remove the HPM and GIM from the MIANet. Then we replace the general class information $p_{gen}$ in the information fusion module with the instance prototype $p$ to establish the baseline. The rest of the experimental settings are consistent with MIANet.

\begin{table*}[htbp]
	\renewcommand\tabcolsep{3.7pt}
	\centering
	\caption{Performance comparison on PASCAL-5$^i$ in terms of mIoU. The \textbf{best} and \underline{second best} results are highlighted with \textbf{bold} and \underline{underline}, respectively.}
	\begin{tabular}{c|ccccccccccc}
		\hline{\tiny {\tiny }}
		& \multicolumn{1}{c|}{}                             & \multicolumn{5}{c|}{1-shot}                                                                                                                                                                                                     & \multicolumn{5}{c}{5-shot}                                                                                                                                                      \\ \cline{3-12} 
		\multirow{-2}{*}{Backbone} & \multicolumn{1}{c|}{\multirow{-2}{*}{Methods}}    & Fold-0                                 & Fold-1                                 & Fold-2                                 & Fold-3                                 & \multicolumn{1}{c|}{Mean}                                   & Fold-0                        & Fold-1                                & Fold-2                        & Fold-3                                 & Mean                          \\ \hline
		& \multicolumn{1}{c|}{PFENet(TPAMI'20)\cite{pfenet}}                       & 56.90                                  & 68.20                                  & 54.40                                  & 52.40                                  & \multicolumn{1}{c|}{58.00}                                  & 59.00                         & 69.10                                  & 54.80                         & 52.90                                  & 59.00                         \\
		& \multicolumn{1}{c|}{HSNet(ICCV'21)\cite{hsnet}}                       & 59.60                                  & 65.70                                  & 59.60                                  & 54.00                                  & \multicolumn{1}{c|}{59.70}                                  & 64.90                         & 69.00                                  & 64.10                         & 58.60                                  & 64.10                         \\
		& \multicolumn{1}{c|}{DPCN(CVPR'22)\cite{dpcn}}                         & 58.90                                  & 69.10                                  & 63.20                                  & 55.70                                  & \multicolumn{1}{c|}{61.70}                                  & 63.40                         & 70.70                                  & 68.10                         & 59.00                                  & 65.30                         \\
		VGG16                      & \multicolumn{1}{c|}{BAM(CVPR'22)\cite{bam}}                          & \underline{63.18}                                  & 70.77                                  & \underline{66.14}                                  & \underline{57.53}                                  & \multicolumn{1}{c|}{\underline{64.41}}                                  &\underline{67.36}                & \underline{73.05}                                  & \underline{70.61}               & \underline{64.00}                                  & \underline{68.76}               \\
		& \multicolumn{1}{c|}{NTRENet(CVPR'22)\cite{ntre}}                         & 57.70                                  & 67.60                                  & 57.10                                  & 53.70                                  & \multicolumn{1}{c|}{59.00}                                  & 60.30                         & 68.00                                  & 55.20                         & 57.10                                  & 60.20                         \\
		& \multicolumn{1}{c|}{Baseline}                     & 56.12                                  & \underline{70.86}                                  & 63.10                                  & 54.36                                  & \multicolumn{1}{c|}{61.11}                                  & 59.92                         & 72.03                                  & 64.69                         & 57.16                                  & 63.45                         \\
		& \multicolumn{1}{c|}{\cellcolor[HTML]{EFEFEF}MIANet} & \cellcolor[HTML]{EFEFEF}\textbf{65.42} & \cellcolor[HTML]{EFEFEF}\textbf{73.58} & \cellcolor[HTML]{EFEFEF}\textbf{67.76} & \cellcolor[HTML]{EFEFEF}\textbf{61.65} & \multicolumn{1}{c|}{\cellcolor[HTML]{EFEFEF}\textbf{67.10}} & \cellcolor[HTML]{EFEFEF}\textbf{69.01} & \cellcolor[HTML]{EFEFEF}\textbf{76.14} & \cellcolor[HTML]{EFEFEF}\textbf{73.24} & \cellcolor[HTML]{EFEFEF}\textbf{69.55} & \cellcolor[HTML]{EFEFEF}\textbf{71.99} \\ \hline \hline
		& \multicolumn{1}{c|}{PFENet(TPAMI'20)\cite{pfenet}}                       & 61.70                                  & 69.50                                  & 55.40                                  & 56.30                                  & \multicolumn{1}{c|}{60.80}                                  & 63.10                         & 70.70                                  & 55.80                         & 57.90                                  & 61.90                         \\
		& \multicolumn{1}{c|}{HSNet(ICCV'21)\cite{hsnet}}                         & 64.30                                  & 70.70                                  & 60.30                         & 60.50                                  & \multicolumn{1}{c|}{64.00}                                  & \underline{70.30}                         & 73.20                                  & 67.40              & 67.10                                  & 69.50                         \\
		& \multicolumn{1}{c|}{DPCN(CVPR'22)\cite{dpcn}}                         & 65.70                                  & 71.60                                  & \textbf{69.10}                         & 60.60                                  & \multicolumn{1}{c|}{66.70}                                  & 70.00                         & 73.20                                  & \underline{70.90}                & 65.50                                  & 69.90                         \\
		ResNet50                   & \multicolumn{1}{c|}{BAM(CVPR'22)\cite{bam}}                          & \textbf{68.97}                         & \underline{73.59}                                  & \underline{67.55}                                  & \underline{61.13}                                  & \multicolumn{1}{c|}{\underline{67.81}}                                  & \textbf{70.59}                & \underline{75.05}                                  & 70.79                         & \underline{67.20}                        & \underline{70.91}                \\
		& \multicolumn{1}{c|}{NTRENet(CVPR'22)\cite{ntre}}                         & 65.40                                  & 72.30                                  & 59.40                                  & 59.80                                  & \multicolumn{1}{c|}{64.20}                                  & 66.20                         & 72.80                                  & 61.70                         & 62.20                                  & 65.70              \\           
		& \multicolumn{1}{c|}{SSP(ECCV'22)\cite{ssp}}                         & 60.50                                  & 67.80                                  & 66.40                         & 51.00                                  & \multicolumn{1}{c|}{61.40}                                  & 67.50                         & 72.30                                  & \textbf{75.20}              & 62.10                                  & 69.30                         \\  
		& \multicolumn{1}{c|}{Baseline}                                          & 61.87                                  & 72.78                                  & 64.10                                  & 55.17                                  & \multicolumn{1}{c|}{63.48}                                                       & 63.36                         & 73.87                                  & 66.50                         & 59.34                                  & 65.77                         \\
		& \multicolumn{1}{c|}{\cellcolor[HTML]{EFEFEF}MIANet}                     & \cellcolor[HTML]{EFEFEF}\underline{68.51}          & \cellcolor[HTML]{EFEFEF}\textbf{75.76} & \cellcolor[HTML]{EFEFEF}67.46          & \cellcolor[HTML]{EFEFEF}\textbf{63.15} & \multicolumn{1}{c|}{\cellcolor[HTML]{EFEFEF}\textbf{68.72}}                      & \cellcolor[HTML]{EFEFEF}70.20 & \cellcolor[HTML]{EFEFEF}\textbf{77.38} & \cellcolor[HTML]{EFEFEF}70.02 & \cellcolor[HTML]{EFEFEF}\textbf{68.77}          & \cellcolor[HTML]{EFEFEF}\textbf{71.59} \\ \hline
	\end{tabular}
	\label{table1}
\end{table*}

\begin{table*}[htbp]
	\renewcommand\tabcolsep{3.7pt}
	\centering
	\caption{Performance comparison on COCO-20$^i$ in terms of mIoU.The \textbf{best} and \underline{second best} results are highlighted with \textbf{bold} and \underline{underline}, respectively.}
	\begin{tabular}{c|c|ccccc|ccccc}
		\hline
		&                              & \multicolumn{5}{c|}{1-shot}                                                                                                                                                                                & \multicolumn{5}{c}{5-shot}                                                                                                                                                               \\ \cline{3-12} 
		\multirow{-2}{*}{Backbone} & \multirow{-2}{*}{Methods}    & Fold-0                                 & Fold-1                                & Fold-2                                 & Fold-3                                 & Mean                                   & Fold-0                       & Fold-1                                 & Fold-2                                 & Fold-3                                 & Mean                          \\ \hline
		& PFENet(TPAMI'20)\cite{pfenet}                       & 35.40                                  & 38.10                                  & 36.80                                  & 34.70                                  & 36.30                                  & 38.20                         & 42.50                                  & 41.80                                  & 38.90                                  & 40.40                         \\
		& DPCN(CVPR'22)\cite{dpcn}                         & 38.50                                  & 43.70                                  & 38.20                                  & 37.70                                  & 39.50                                  & 42.70                         & 51.60                                  & 45.70                                  & 44.60                                  & 46.20                         \\
		VGG16                      & BAM(CVPR'22)\cite{bam}                          & \underline{38.96}                                  & \underline{47.04}                                  & \underline{46.41}                                  & \underline{41.57}                                  & \underline{43.50}                                  & \textbf{47.02}                & \underline{52.62}                                  & \underline{48.59}                                  & \underline{49.11}                                  & \underline{49.34}                \\ 
		& Baseline                     & 33.55                                  & 41.45                                  & 35.49                                  & 34.46                                  & 36.24                                  & 38.11                         & 49.57                                  & 41.94                                  & 41.53                                  & 42.79                         \\
		& \cellcolor[HTML]{EFEFEF}MIANet & \cellcolor[HTML]{EFEFEF}\textbf{40.56} & \cellcolor[HTML]{EFEFEF}\textbf{50.53} & \cellcolor[HTML]{EFEFEF}\textbf{46.50} & \cellcolor[HTML]{EFEFEF}\textbf{45.18} & \cellcolor[HTML]{EFEFEF}\textbf{45.69} & \cellcolor[HTML]{EFEFEF}\underline{46.18} & \cellcolor[HTML]{EFEFEF}\textbf{56.09} & \cellcolor[HTML]{EFEFEF}\textbf{52.33} & \cellcolor[HTML]{EFEFEF}\textbf{49.54} & \cellcolor[HTML]{EFEFEF}\textbf{51.03} \\ \hline \hline
		& \multicolumn{1}{c|}{HSNet(ICCV'21)\cite{hsnet}}                       & 36.30                                  & 43.10                                  & 38.70                                  & 38.70                                  & \multicolumn{1}{c|}{39.20}                                  & 43.30                         & 51.30                                  & 48.20                         & 45.00                                  & 46.90                         \\
		& DPCN(CVPR'22)\cite{dpcn}                         & 42.00                                  & 47.00                                  & 43.20                                  & 39.70                                  & 43.00                                  & 46.00                         & 54.90                                  & 50.80                                  & 47.40                                  & 49.80                         \\
		ResNet50                   & BAM(CVPR'22)\cite{bam}                          & \textbf{43.41}                         & \underline{50.59}                                 & \underline{47.49}                                  & \underline{43.42}                                  & \underline{46.23}                                  & \textbf{49.26}                & \underline{54.20}                                  & \textbf{51.63}                         & \underline{49.55}                         & \underline{51.16}                \\
		& NTRENet(CVPR'22)\cite{ntre}                      & 36.80                                  & 42.60                                  & 39.90                                  & 37.90                                  & 39.30                                  & 38.20                         & 44.10                                  & 40.40                                  & 38.40                                  & 40.30                         \\
		& \multicolumn{1}{c|}{SSP(ECCV'22)\cite{ssp}}                         & 35.50                                  & 39.60                                  & 37.90                         & 36.70                                  & \multicolumn{1}{c|}{37.40}                                  & 40.60                         & 47.00                                  & 45.10              & 43.90                                  & 44.10 \\  
		& Baseline                     & 36.07                                  & 43.97                                  & 40.23                                  & 39.34                                  & 39.90                                  & 42.79                         & 49.42                                  & 47.41                                  & 46.08                                  & 46.43                          \\
		& \cellcolor[HTML]{EFEFEF}MIANet & \cellcolor[HTML]{EFEFEF}\underline{42.49}          & \cellcolor[HTML]{EFEFEF}\textbf{52.95} & \cellcolor[HTML]{EFEFEF}\textbf{47.77} & \cellcolor[HTML]{EFEFEF}\textbf{47.42} & \cellcolor[HTML]{EFEFEF}\textbf{47.66} & \cellcolor[HTML]{EFEFEF}\underline{45.84} & \cellcolor[HTML]{EFEFEF}\textbf{58.18} & \cellcolor[HTML]{EFEFEF}\underline{51.29}          & \cellcolor[HTML]{EFEFEF}\textbf{51.90}          & \cellcolor[HTML]{EFEFEF}\textbf{51.65} \\ \hline
	\end{tabular}
	\label{table2}
\end{table*}

\subsection{Comparison with State-of-the-Arts}
\noindent\textbf{PASCAL-5$^i$.} Table \ref{table1} shows the mIoU performance comparison on PASCAL-5$^i$ between our method and several representative models. It can be seen that (1) MIANet achieves state-of-the-art performance under the 1-shot and 5-shot settings. Especially for the VGG16 \cite{vgg} backbone, we surpass BAM \cite{bam}, which holds the previous state-of-the-art results, by 2.69\% and 3.23\%. (2) MIANet outperforms the baseline with a large margin. For example, when VGG16 is the backbone, MIANet and the baseline model achieve 67.10\% and 61.11\% respectively. Compared with ResNet50 \cite{resnet}, VGG16 provides less information that is useful for segmentation, so the extra information is more valuable. After adding the detailed general and instance information generated by the GIM and HPM to the baseline model, better performance improvement occurs than ResNet50.\par

\noindent\textbf{COCO-20$^i$.} COCO-20$^i$ is a more challenging dataset that contains multiple objects and shows greater variance. Table \ref{table2} shows the mIoU performance comparison. Overall, MIANet surpasses all the previous methods under 1-shot and 5-shot settings. Under the 1-shot setting, MIANet leads BAM by 2.19\% and 1.43\% on VGG16 and ResNet50. Meanwhile, our method outperforms the baseline by 9.45\%, and 7.76\%, which demonstrate the superiority of our method, despite the challenging scenarios.\par

\begin{figure*}[htbp]
	\centering
	\includegraphics[width=1\linewidth,height=0.4\linewidth]{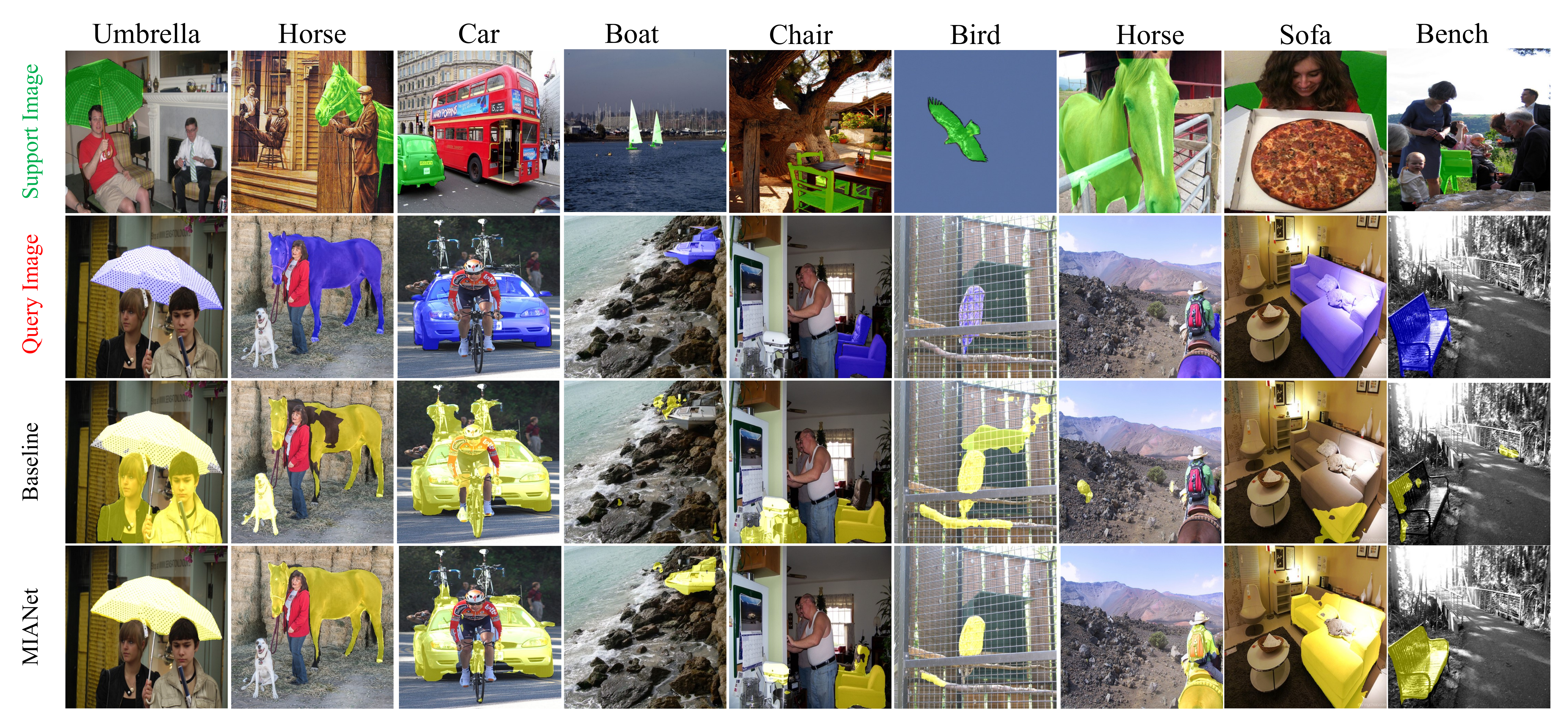}
	\caption{Qualitative results of our method MIANet and baseline on PASCAL-5$^i$ and COCO-20$^i$ benchmarks. Zoom in for details.}
	\label{figure4}
\end{figure*}

\noindent\textbf{Qualitative Results.} We report some qualitative results generated from our MIANet and baseline model on the PASCAL-5$^i$ and COCO-20$^i$ benchmarks. Compared with the baseline, MIANet exhibits the following advantages as shown in Figure \ref{figure4}. (1) MIANet can more accurately segment the target class, while the baseline incorrectly segments the seen classes as the target classes (1st to 3rd columns). (2) MIANet can mine similar local features for different fine-grained categories to address the intra-class variance problem caused by semantic differences, i.e., sailboat/small boat, chair/sofa chair, and eagle/owl in the 4th, 5th and 6th columns respectively. (3) MIANet can provide general information that is missing in the support image (7th to 9th columns), i.e., the intra-class variance caused by perspective distortion.

\subsection{Ablation study}
We conduct extensive ablation studies on PASCAL-5$^i$ under the 1-shot setting to validate the effectiveness of our proposed key modules, i.e., HPM, and GIM. Note that the experiments in this section are performed on PASCAL-5$^i$ dataset using VGG16 backbone. Moreover, we provide experiment details and extra experiments in \textbf{Supplementary Materials}. \par
\noindent\textbf{Components Analysis.} Table \ref{table3} shows the impact of each component on the model performance. Overall, using the two components proposed in this paper improves the baseline by 5.99\%. In the second row, HPM mines the multi-scale instance-level information and improves the baseline by 3.44\%. Meanwhile, replacing the support prototype $p$ with the general prototype $p_{gen}$, the baseline yields a 1.35\% performance gain. This is because GIM produces general information, while HPM can discover pixel-level information of instances, which is more helpful for the improvement of segmentation performance. After the combination of GIM and HPM, the instance information and general information are aggregated by IFM so that the model can alleviate the problem of intra-class differences, and effectively improve the performance by 2.55\% compared to the second row.\par

\begin{table}[]
	\centering
	\renewcommand\tabcolsep{2.2pt}
	\caption{Ablation studies of main model components.}
	\begin{tabular}{ll|llll|l}
		\Xhline{1.5pt}
		HPM & GIM & Fold-0 & Fold-1 & Fold-2 & Fold-3 & mIoU \\ \Xhline{1pt}
		     &     & 56.12 & 70.86 & 63.10 & 54.36 & 61.11 \\
		   \checkmark  &     & 61.58 & 71.80 & 67.06 &57.75 & 64.55$_{\uparrow 3.44}$ \\
		    & \checkmark    & 61.02 & 72.11 & 63.77 & 52.95 & 62.46$_{\uparrow 1.35}$ \\
		  \checkmark   &  \checkmark   & 65.42 & 73.58 & 67.76 & 61.65 & 67.10$_{\uparrow 5.99}$ \\ \Xhline{1.5pt}
	\end{tabular}
	\label{table3}
\end{table}

\noindent\textbf{Hierarchical Prior Module.} HPM uses multi-scale prior information and establishes information channels with weighted average pooling between different scales, which provides instance-level prior information for MIANet. Table \ref{table4} shows the impact of each element in HPM on the model performance. We can see that using the proposed multi-scale prior outperforms the one-scale method by 1.69\%. This is because multi-scale instance information can adapt to input objects of different sizes. In addition, by establishing information paths between different scales, the proposed weighted pooling method can also avoid losing discriminative features and achieve a performance improvement of 0.48\%.\par

\begin{table}[]
	\centering
	\renewcommand\tabcolsep{2.2pt}
	\caption{Ablation studies of the main elements in HPM. The baseline is equipped with GIM. "OS" means the HPM employs the one-scale prior information, "MS" means the multi-scale method, and "IC" denotes the information channels.}
	\begin{tabular}{lll|llll|l}
		\Xhline{1.5pt}
		OS &MS & IC & Fold-0 & Fold-1 & Fold-2 & Fold-3 & mIoU \\ \Xhline{1pt}
		   &  &     & 61.02 & 72.11 & 63.77 & 52.95 & 62.46 \\
		   \checkmark&  &     & 64.08 & 72.40 & 65.27 & 57.97 & 64.93$_{\uparrow 2.47}$ \\
		   &  \checkmark &     & 64.52 &73.07 & 67.75 & 61.13 & 66.62$_{\uparrow 4.16}$ \\
		  &\checkmark   &  \checkmark   & 65.42 & 73.58 & 67.76 & 61.65 & 67.10$_{\uparrow 4.64}$ \\ \Xhline{1.5pt}
	\end{tabular}
	\label{table4}
\end{table}

\noindent\textbf{General Information Module.} Table \ref{table5} shows the impact of main components in GIM, namely triplet loss, and word embeddings. After removing the triplet loss, the performance drops by 0.61\%. This is because the triplet loss pulls together similar local features and pushes away dissimilar ones in $l_2$ metric space, and learns better general information representations for MIANet. Second, when we directly remove the word embedding in Figure \ref{figure3} and only use the instance class prototype as the input of the general information generator, the performance drops by 1.34\%.

\begin{table}[]
	\centering
	\renewcommand\tabcolsep{3pt}
	\caption{Ablation studies of main components in GIM. The baseline is equipped with HPM. "TL" and "WE" denotes the triplet loss and word embeddings respectively. }
	\begin{tabular}{ll|llll|l}
		\Xhline{1.5pt}
		TL & WE & Fold-0 & Fold-1 & Fold-2 & Fold-3 & mIoU \\ \Xhline{1pt}
		\checkmark   &  \checkmark   & 65.42 & 73.58 & 67.76 & 61.65 & 67.10 \\ 
		  &  \checkmark   & 63.99 &73.09 & 67.65 & 61.22 & 66.49$_{\downarrow 0.61}$ \\
		  \checkmark &     & 63.64 & 71.47 & 67.72 & 60.20 & 65.76$_{\downarrow 1.34}$ \\ 
		    &     &  61.58 & 71.80 & 67.06 &57.75 & 64.55$_{\downarrow 2.55}$ \\ \Xhline{1.5pt}
		
	\end{tabular}
	\label{table5}
\end{table}

\section{Conclusion}
We propose a multi-information aggregation network (MIANet) with three major parts (i.e., HPM, GIM and IFM) for the few-shot semantic segmentation. The non-parametric HPM generates unbiased multi-scale instance information at the pixel level while alleviating the prediction bias problem of the model. The GIM obtains additional general class prototypes from word embeddings, as a supplement to the instance information. A triplet loss is designed to optimize the GIM to make the prototypes better alleviate the intra-class variance problem. The instance-level information and general information are aggregated in IFM, which is beneficial to more accurate segmentation results. Comprehensive experiments show that MIANet achieves state-of-the-art performance under all settings.

{\small
\bibliographystyle{ieee_fullname}
\bibliography{egbib}
}

\clearpage
\appendix
\section{Appendix}
\subsection{Implement details}
\begin{itemize}[itemsep=1pt,topsep=0pt,parsep=0pt]
	\item [(1)] In the hierarchical prior module (HPM) of MIANet, the size of $M_{ins}$ is $\left\{(60,60), (30, 30), (15, 15), (8, 8)\right\}$, which is consistent with PFENet \cite{pfenet}.
	\item [(2)] In the general information module (GIM), the \textit{middle-level features} are obtained by concatenating the intermediate features of backbone. For instance, we get the middle-level features of ResNet50 through concatenating the features from block 2 and block 3 \cite{canet}. The middle-level feature dimension $c$ is 256.
\end{itemize}

\subsection{Comparison with State-of-the-art Methods}
First, we list the FB-IoU results in Table \ref{sup_table1}, where the proposed method can gain great improvement, especially in the case of using the VGG16.\par Then we report the results in Table \ref{sup_table2} when the ResNet101 is used as the backbone under 1-shot settings. It can be seen that our approach achieves new state-of-the-art performance and outperforms previous state-of-the-art result by 1.43\%.
\begin{table}[htbp]
	\centering
	\renewcommand\tabcolsep{2pt}
	\caption{Performance comparison in terms of FB-IoU. The results are the averaged FB-IoU scores of all the four folds. "VGG" means the backbone of VGG16, and "ResNet" means ResNet50.}
	\begin{tabular}{c|c|cc|cc}
		\hline
		&                                & \multicolumn{2}{c|}{1-shot}                                                     & \multicolumn{2}{c}{5-shot}                                                      \\ \cline{3-6} 
		\multirow{-2}{*}{Datasets} & \multirow{-2}{*}{Methods}      & VGG                                  & ResNet                               & VGG                                  & ResNet                               \\ \hline
		& PFENet\cite{pfenet}                         & 72.00                                  & 73.30                                  & 72.30                                  & 73.90                                  \\
		& HSNet\cite{hsnet}                          & 73.40                                  & 76.70                                  & 76.60                                  & 80.60                                  \\
		& DPCN\cite{dpcn}                          & 73.70                                  & 78.00                                  & 77.20                                  & 80.70                                  \\
		PASCAL-$5^i$                     & BAM\cite{bam}                            & 77.26                                  & \textbf{81.10}                         & 79.71                                  & 82.18                                  \\
		& NTRENet\cite{ntre}                        & 73.10                                  & 77.00                                  & 74.20                                  & 78.40                                  \\ \cline{2-6} 
		& \cellcolor[HTML]{EFEFEF}MIANet & \cellcolor[HTML]{EFEFEF}\textbf{79.22} & \cellcolor[HTML]{EFEFEF}79.54          & \cellcolor[HTML]{EFEFEF}\textbf{82.69} & \cellcolor[HTML]{EFEFEF}\textbf{82.20} \\ \hline
		& HSNet\cite{hsnet}                          & -                                      & 68.20                                  & -                                      & 70.70                                  \\
		COCO-$20^i$                         & DPCN\cite{dpcn}                           & 62.50                                  & 63.20                                  & 66.10                                  & 67.40                                  \\
		& NTRENet\cite{ntre}                        & -                                      & 68.50                                  & -                                      & 69.20                                  \\ \cline{2-6} 
		& \cellcolor[HTML]{EFEFEF}MIANet & \cellcolor[HTML]{EFEFEF}\textbf{71.01} & \cellcolor[HTML]{EFEFEF}\textbf{71.51} & \cellcolor[HTML]{EFEFEF}\textbf{73.81} & \cellcolor[HTML]{EFEFEF}\textbf{73.13} \\ \hline
	\end{tabular}
	\label{sup_table1}
\end{table}

\begin{table}[]
	\centering
	\renewcommand\tabcolsep{3.5pt}
	\caption{Performance comparison on PASCAL-$5^i$ when using ResNet101.}
	\begin{tabular}{c|cccc|l}
		\Xhline{1.5pt}
		Margin & Fold-0          & Fold-1          & Fold-2          & Fold-3          & mIoU           \\ \Xhline{1pt}
		PFENet\cite{pfenet}    & 60.50          & 69.40 & 54.40         & 55.90          & 60.10          \\
		HSNet\cite{hsnet}      & 67.30          & 72.30         & 62.00 & \textbf{63.10} & 66.20          \\
		NTRENet\cite{ntre}  &65.50 & 71.80          & 59.10          & 58.30          & 63.70 \\
		\cellcolor[HTML]{EFEFEF}MIANet      & \cellcolor[HTML]{EFEFEF}\textbf{68.54}          & \cellcolor[HTML]{EFEFEF}\textbf{76.34}         & \cellcolor[HTML]{EFEFEF}\textbf{64.92}         &\cellcolor[HTML]{EFEFEF} 60.70          &\cellcolor[HTML]{EFEFEF} \cellcolor[HTML]{EFEFEF}\textbf{67.63}         \\ \Xhline{1.5pt}
	\end{tabular}
	\label{sup_table2}
\end{table}

\subsection{Ablation study}
We conduct extra ablation studies to validate the impact of our designs. Note that the experiments in this section are performed on PASCAL-5$^i$ dataset using the VGG16 backbone unless specified otherwise. And the evaluation metric is mean-IoU.\par

\vspace{2mm}
\noindent\textbf{Effect of the averaging strategy.} In MIANet, we average the negative set since the elements in the background of the support images are very complex. We show the result in Table \ref{sup_table4} if the averaging strategy is not implemented. Averaging the background elements brings a 1\% performance gain.

\begin{table}[]
	\centering
	\renewcommand\tabcolsep{3pt}
	\caption{Ablation studies of the averaging strategy.}
	\begin{tabular}{c|llll|l}
		\Xhline{1.5pt}
		Average & Fold-0 & Fold-1 & Fold-2 & Fold-3 & mIoU \\ \Xhline{1pt}
		& 63.84 & 72.75 & 67.44 & 60.38 & 66.10 \\
		\checkmark   & 65.42 & 73.58 & 67.76 & 61.65 & \textbf{67.10}\\
		\Xhline{1.5pt}
	\end{tabular}
	\label{sup_table4}
\end{table}

\vspace{2mm}
\noindent\textbf{Effect of the pretrained strategy.} Current s-o-t-a methods \cite{bam,cwt} usually adopt the pretrained strategy to pretrain the backbone before meta-training. We conduct the experiment in Table \ref{sup_table5} which demonstrates the effectiveness of the strategy.

\begin{table}[]
	\centering
	\renewcommand\tabcolsep{3pt}
	\caption{Ablation studies of the pretrained strategy.}
	\begin{tabular}{c|llll|l}
		\Xhline{1.5pt}
		Pretrained & Fold-0 & Fold-1 & Fold-2 & Fold-3 & mIoU \\ \Xhline{1pt}
		&63.56 & 72.92 & 65.48 &58.18 & 65.03 \\
		\checkmark   & 65.42 & 73.58 & 67.76 & 61.65 & \textbf{67.10}\\
		\Xhline{1.5pt}
	\end{tabular}
	\label{sup_table5}
\end{table}
\vspace{2mm}
\noindent\textbf{Effect of the margin.} We report the ablation study about how to choose the margin in our proposed triplet loss, whose results are listed in \ref{sup_table6}. The best result is achieved when the margin is 0.5.

\begin{table}[]
	\centering
	\caption{Ablation studies of the margin in triplet loss on PASCAL-$5^i$when using ResNet50.}
	\begin{tabular}{c|cccc|l}
		\Xhline{1.5pt}
		Margin & Fold-0          & Fold-1          & Fold-2          & Fold-3          & mIoU           \\ \Xhline{1pt}
		0.1    & 67.69          & \textbf{76.30} & 67.09          & 61.84          & 68.23          \\
		0.2    & 66.75          & 75.32          & \textbf{67.82} & \textbf{63.20} & 68.27          \\
		0.5    & \textbf{68.51} & 75.76          & 67.46          & 63.15          & \textbf{68.72} \\
		1      & 68.32          & 75.23          & 66.72          & 62.47          & 68.19          \\ \Xhline{1.5pt}
	\end{tabular}
	\label{sup_table6}
\end{table}

\begin{figure}[htbp]
	\centering
	\includegraphics[width=1\linewidth,height=0.5\linewidth]{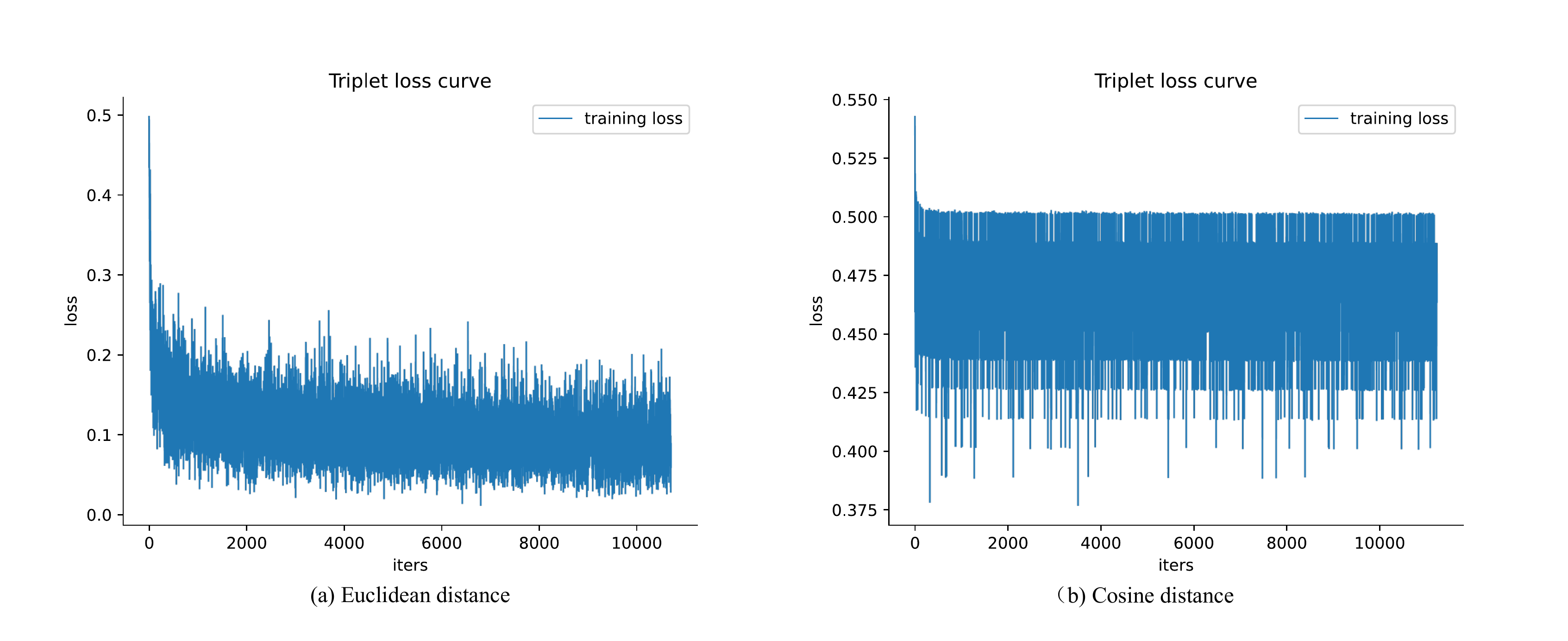}
	\caption{Visual display of the triplet loss in training when using different metric tools.}
	\label{sup_figure4}
\end{figure}

\vspace{2mm}
\noindent\textbf{Effect of the metric tools in the triplet loss.} In the triplet loss, euclidean distance is used as our metric tool to calculate the distance of triplets. We investigate two types of metric tools, i.e. euclidean distance and cosine distance. The results are listed in Table \ref{sup_table7}. The euclidean distance leads the performance by 1.96\%. As Figure \ref{sup_figure4} shows, euclidean distance makes MIANet learn better from the hard triplets. When using the cosine distance, the value of the triplet loss is maintained around 0.5 (\textbf{margin}), which means that the triplet loss cannot distinguish the positive samples and negative samples well.

\begin{table}[]
	\centering
	\renewcommand\tabcolsep{2pt}
	\caption{Ablation studies of the metric tools.}
	\begin{tabular}{c|llll|l}
		\Xhline{1.5pt}
		Methods & Fold-0 & Fold-1 & Fold-2 & Fold-3 & mIoU \\ \Xhline{1pt}
		cosine distance &62.65  &72.51  &68.72 &56.67 & 65.14 \\
		euclidean distance   & 65.42 & 73.58 & 67.76 & 61.65 & \textbf{67.10}\\
		\Xhline{1.5pt}
	\end{tabular}
	\label{sup_table7}
\end{table}

\subsection{More Visualizations}

\begin{figure*}[htbp]
	\centering
	\includegraphics[width=1\linewidth,height=0.5\linewidth]{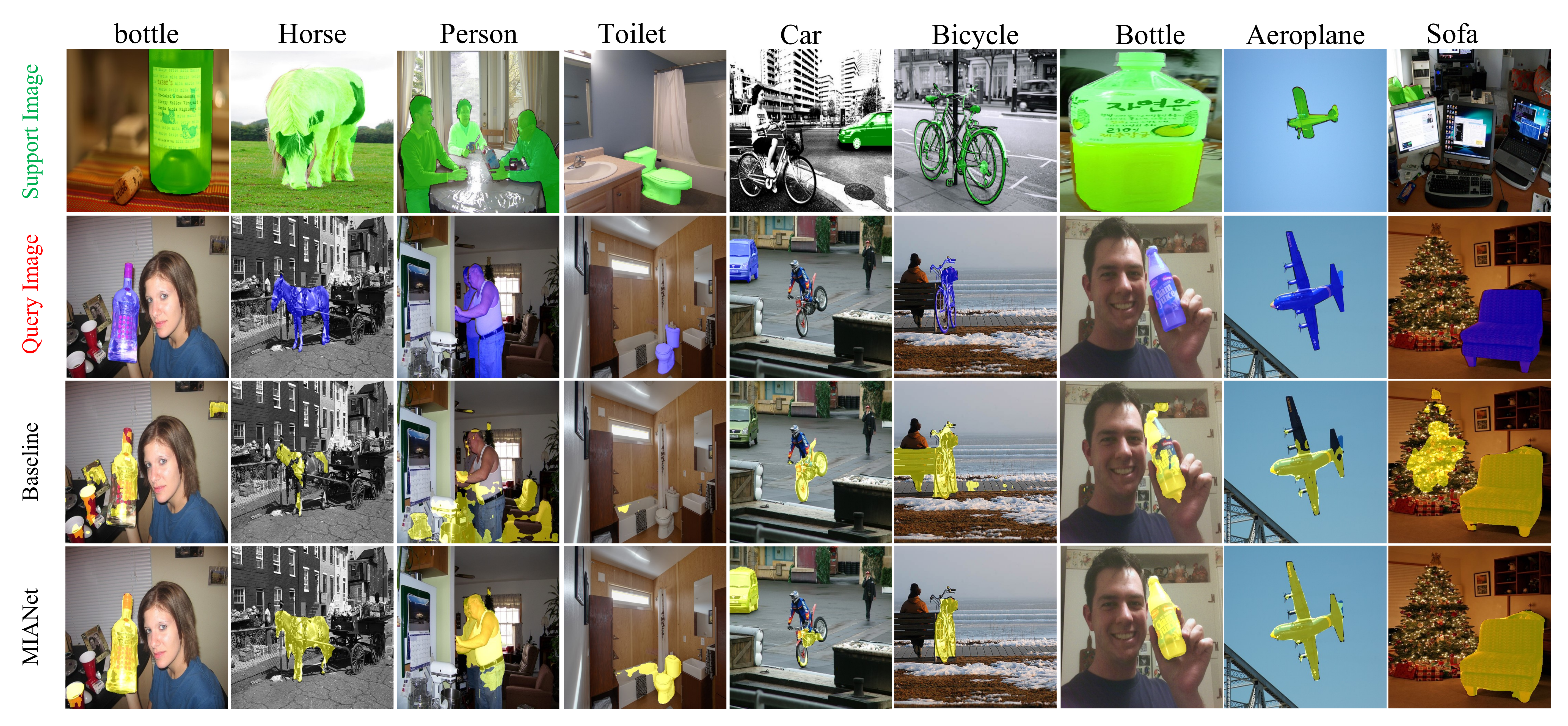}
	\caption{Qualitative results of our method MIANet and baseline on PASCAL-5$^i$ and COCO-20$^i$ benchmarks. Zoom in for details.}
	\label{sup_figure2}
\end{figure*}

\begin{figure*}[htbp]
	\centering
	\includegraphics[width=1\linewidth,height=0.5\linewidth]{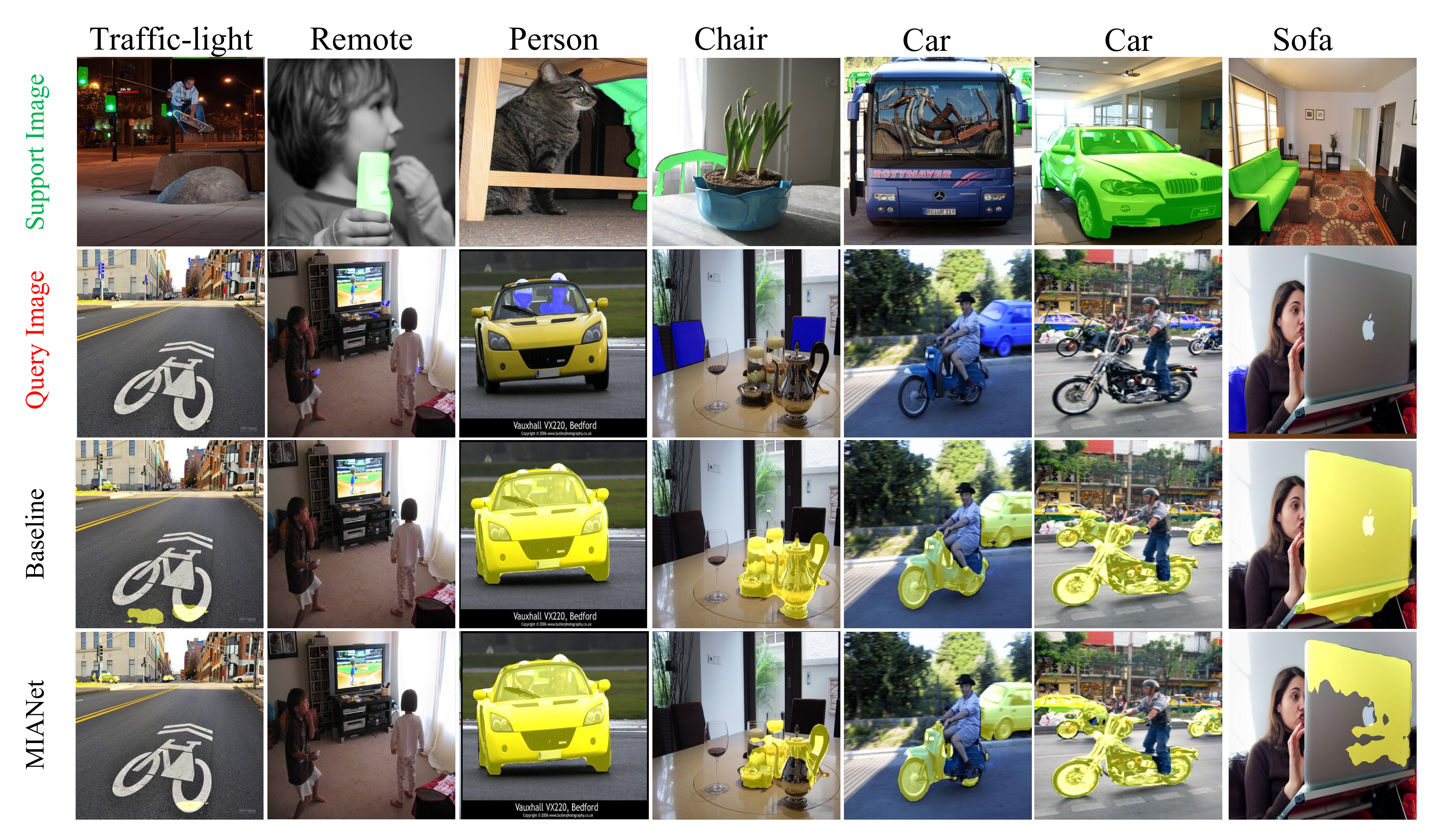}
	\caption{\textbf{Failure results} of our method MIANet and baseline on PASCAL-5$^i$ and COCO-20$^i$ benchmarks. Zoom in for details.}
	\label{sup_figure3}
\end{figure*}

\vspace{2mm}
We demonstrate more qualitative results in Figure \ref{sup_figure2}. Moreover, some \textbf{failure cases} are also provided in Figure \ref{sup_figure3}. As the Figure \ref{sup_figure3} shows, we can conclude that (1) intra-class differences seriously affect the segmentation performance, especially the cases of perspective distortion (2nd, 3rd, and 7th columns). (2) The segmentation of small objects is also unsatisfactory (1st and 2nd columns). (3) The bias to the base classes is still an urgent problem in few-shot segmentation (5th and 6th columns). How to more effectively deal with these problems requires better modeling of changes in views, pose and occlusion.

\end{document}